\documentclass[letterpaper, 10 pt, conference]{ieeeconf}  

\IEEEoverridecommandlockouts                              

\usepackage{graphics} 
\usepackage{graphicx} 
\usepackage{amsmath} 
\usepackage{amssymb}  
\usepackage{xcolor}
\usepackage{lipsum}
\usepackage{soul}
\usepackage{hyperref}
\usepackage{subcaption}
\usepackage[ruled,vlined,linesnumbered,noend]{algorithm2e}
\usepackage{comment}
\usepackage{tabularx}
\usepackage{siunitx}
\usepackage{wrapfig}
\usepackage[noadjust,compress]{cite}

\newtheorem{theorem}{Theorem}

\usepackage{xspace}

\newcommand{\name}{{\tt KRAFT}}

\DeclareMathOperator*{\argmin}{arg\,min}

\newenvironment{myitem}
{
    \begin{list}{$\bullet$ }{}
        \setlength{\topsep}{0pt}
        \setlength{\parskip}{0pt}
        \setlength{\labelsep}{0pt}
        \setlength{\partopsep}{0pt}
        \setlength{\parsep}{0pt}         
        \setlength{\itemsep}{0pt} 
	    \setlength{\leftskip}{-20pt}
        \setlength{\leftmargin}{0pt}
}
{
    \end{list} 
}

\title{\LARGE \bf \name: Sampling-Based Kinodynamic Replanning and Feedback Control over Approximate, Identified Models of Vehicular Systems}

\author{Aravind Sivaramakrishnan, Sumanth Tangirala, Dhruv Metha Ramesh, Edgar Granados, and Kostas E. Bekris
\thanks{The authors are with the Dept. of Computer Science, Rutgers University, NJ, USA. E-mail: {\tt \{as2578, kb572\}@rutgers.edu}.}%
}

\begin{document}

\maketitle
\thispagestyle{empty}
\pagestyle{empty}

\begin{abstract}

This paper aims to increase the safety and reliability of executing trajectories planned for robots with non-trivial dynamics given a light-weight, approximate dynamics model. Scenarios include mobile robots navigating through workspaces with imperfectly modeled surfaces and unknown friction.  The proposed approach, 
{\it K}inodynamic {\it R}eplanning over {\it A}pproximate Models with {\it F}eedback {\it T}racking (\name), integrates: (i) replanning via an asymptotically optimal sampling-based kinodynamic tree planner, with (ii) trajectory following via feedback control, and (iii) a safety mechanism to reduce collision due to second-order dynamics. The planning and control components use a rough dynamics model expressed analytically via differential equations, which is tuned via system identification (SysId) in a training environment but not the deployed one. This allows the process to be fast and achieve long-horizon reasoning during each replanning cycle. At the same time, the model still includes gaps with reality, even after SysID, in new environments.  Experiments demonstrate the limitations of kinematic path planning and path tracking approaches, highlighting the importance of: (a) closing the feedback-loop also at the planning level; and (b) long-horizon reasoning, for safe and efficient trajectory execution given inaccurate models. 
Website: \url{https://prx-kinodynamic.github.io/projects/kraft}


\end{abstract}


\section{Introduction}
\label{sec:introduction}

This work aims to improve the safety and efficiency of executing trajectories for robots with non-trivial dynamics planned using approximate models. Consider a mobile robot, especially a low-cost one, navigating an environment that involves imperfectly modeled surfaces, e.g., speed bumps, ramps, and unknown friction. The robot has access to an approximate dynamics model for planning, which does not reflect the true dynamics upon execution. The approximate model may be tuned in a training environment, but the robot is deployed in different locations, each with varying properties. Then, open-loop execution of trajectories leads to significant deviations and collisions due to the model gap.  Given (imperfect) state estimation, various controllers have been proposed for tracking planned paths \cite{conlter1992implementation,quinlan1993elastic, fox1997dynamic} or trajectories \cite{hoffmann2007autonomous}, and their application is often considered sufficient for safety, e.g., a traditional solution is to plan a kinematic path and employ a path follower. The accompanying experiments, however, show that this approach fails to provide safety for the considered challenges. Planning a kinodynamically feasible trajectory and adopting a trajectory follower reduces deviations but the accompanying experiments show that it still leads to failures even after significant tuning, especially when effects like non-flat terrain cause the robot to significantly deviate from its plan. 

In this context, this work proposes \textbf{K}inodynamic \textbf{R}eplanning with \textbf{A}pproximate Models and \textbf{F}eedback \textbf{T}racking (\name), which performs online replanning using a sampling-based kinodynamic tree planner and uses a trajectory follower to increase the reliability of successful execution given model gaps. The approach follows the principles of model-predictive control (MPC), which re-computes trajectories given the latest state predictions. Given a conservative estimate of the model gap, the approach also allows to incorporate \textit{contingency} plans to ensure safety. These trajectories are passed down to a trajectory follower, which operates at a high frequency for tracking, while also reasoning about the robot's dynamics. 

Most MPC solutions \cite{park2016graceful, williams2017information, 8558663} reason over a short, finite horizon with strategies, such as selecting a short-term feasible maneuver that optimizes a cost map that requires tuning to return desirable solutions. In contrast, the proposed approach adopts a sampling-based kinodynamic planner, which reasons over longer horizons. This leads to more efficient solutions (in terms of execution time) and requires less parameter tuning beyond access to a dynamics model and a cost function. Overall, \name\ has the following features: 

\begin{figure}[t]
    \centering
    \includegraphics[width=.4925\columnwidth]{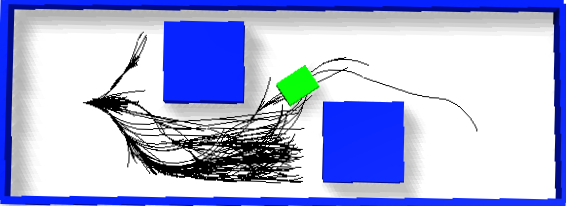}
    \includegraphics[width=.4925\columnwidth]{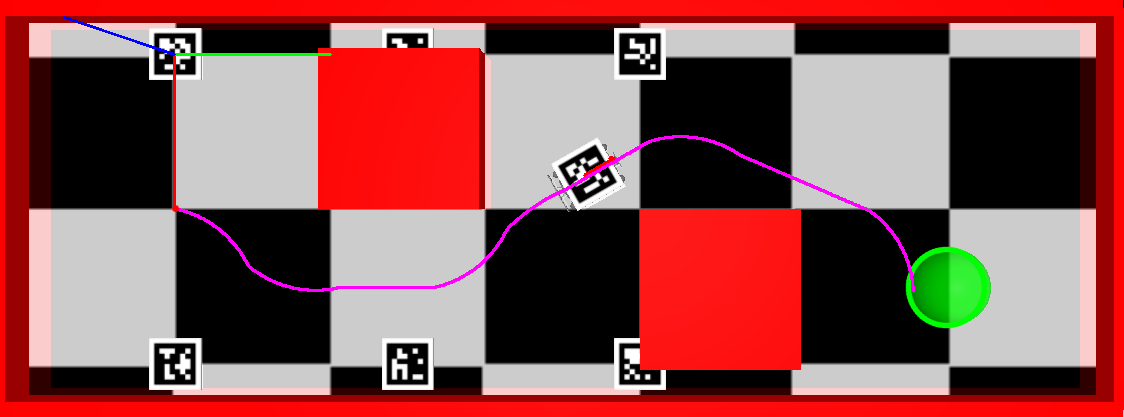}\\
    \vspace{0.05in}
    \includegraphics[width=.9\columnwidth]{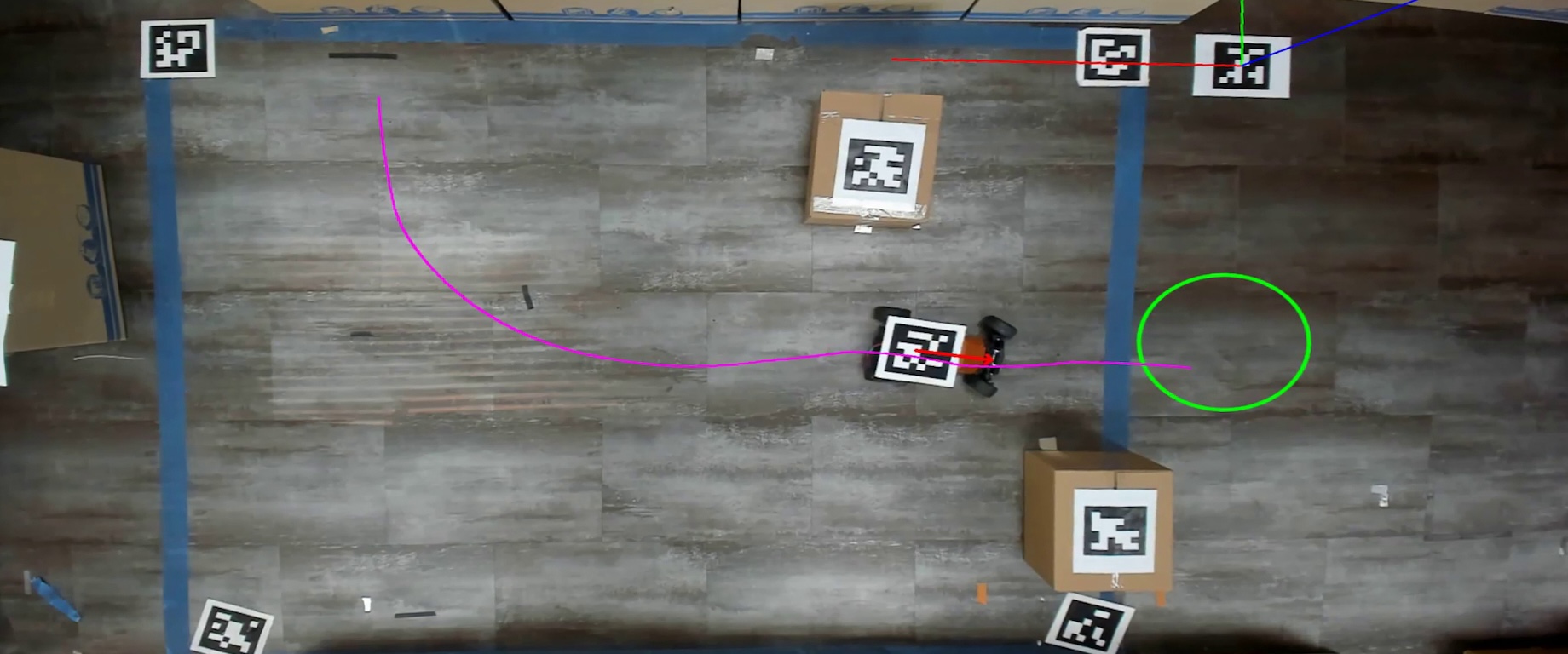}
    \caption{(Top-Left) The initial planned tree given an analytical model. (Top-Right) Execution in MuJoCo integrating kinodynamic replanning and trajectory following. (Bottom) A similar experiment with a real MuSHR.}
    \label{fig:intro-figure}
    \vspace{-.15in}
\end{figure}




\begin{myitem}
    \item Employs low-fidelity but tunable dynamics models that allow fast reasoning, which are identified in a standard environment but not necessarily tuned for the deployed one;
    \item Online replanning with an asymptotically optimal and informed  sampling-based kinodynamic tree planner that uses the low-fidelity dynamics model and robot state predictions;
    \item Trajectory following via feedback control given the same dynamics model and the latest robot state observations;
    \item A safety framework to minimize in a computationally efficient manner undesirable collisions due to the model gap.
\end{myitem}
The accompanying experiments demonstrate the effectiveness of \name\  first on physically-simulated benchmarks, where the planner's dynamics model is analytical. It also demonstrates the proposed framework on a real, low-cost robotic platform, the MuSHR \cite{srinivasa2019mushr}, on a navigation task.

\section{Related Work}
\label{sec:prelims}

\textbf{Sampling-based Motion Planners (SBMPs)} can be used for kinodynamic planning and provide desirable properties, such as Asymptotic Optimality (AO), where recent progress has been achieved on acquiring high-quality solutions fast \cite{LB-DIRT, chiang2019rl, sivaramakrishnan2021improving, troy2022terrains, sivaramakrishnan2023roadmaps}. To improve executability of solutions, sampling-based feedback planners \cite{tedrake2010lqr, reist2016feedback, majumdar2017funnel, m2023pip} consider controllers during the planning process to ensure trajectory tracking under model uncertainty. They tend to be computationally demanding, however, even for simple models and would have to be built specifically for the deployed environment.

\textbf{Replanning with sampling-based planners} is a powerful tool for generating robot motions in real time under disturbances \cite{tsianos2008replanning}. It was first demonstrated in environments with dynamic obstacles \cite{kindel2000} but safety issues related to Inevitable Collision States (ICS) \cite{fraichard2004inevitable} arise when replanning with significant dynamics. It is possible, however, to ensure the existence of safe contingency maneuvers (e.g., braking) at the every planning cycle while minimizing the cost of collision-checking \cite{bekris2007greedy}. Adapting planning cycle duration also allows a tradeoff between safety and path quality \cite{hauser2010adaptive}. These works typically assume a perfect execution model. The current paper aims to address this via integration with system identification and feedback control, while utilizing the progress achieved by sampling-based motion planning.

\textbf{Planning with Inaccurate Models} has also been approached with search-based methods \cite{Vemula-RSS-20,vemula2021cmax++}, where a penalty term is introduced for state transitions where the model was observed to be inaccurate during execution. When a controller is used to track a planned path, a similar approach \cite{ratner2023inaccurate} introduced a \textit{control-level discrepancy model} that biases the search away from transitions the controller cannot track reliably. \textit{Model Deviation Estimate} (MDE) \cite{mcconachie2020learning,mitrano2021learning} biases an SBMP away from regions where the deviation is expected to be high. Training this estimator involves collecting data in the environment where the robot is deployed. The current paper does not assume new execution data in the deployment environment but does allow system identification in a training environment. 

{\bf System identification} estimates parameters such that the simulated model minimizes deviations from the real robot. Methods include convex optimization \cite{wang2019convex,lee2019geometric}, Koopman operators \cite{bruder2019nonlinear}, factor graphs \cite{burri2018framework} and machine learning \cite{wehbe2017experimental}. Differentiable physics simulators (DPS) have been used to identify system parameters, ensuring stable performance of both omnidirectional \cite{granados2022model} and Ackermann-steered vehicles \cite{gonultas2023system}. Closed-box neural networks \cite{atreya2022high} that have also been integrated with non-linear least squares optimization for path following.  This work opts for a \textit{factor graphs}-based dynamics model \cite{dellaert2021factor} whose parameters can be identified with relatively low data requirements. 





\section{Problem Statement}

Consider a mobile robot with state space $\mathbb{X}$ and control space $\mathbb{U}$ navigating a workspace $\mathbb{W}$ from an initial state $x_s$ to a goal region $X_G$. The robot has a map of known, static obstacles it should not collide with, which divides $\mathbb{X}$ into collision-free $\mathbb{X}_\mathrm{f}$ and obstacle $\mathbb{X}_\mathrm{o}$ subsets. The {\bf true dynamics} $\dot{x} = f(x,u)$, $x \in \mathbb{X}$, $u \in \mathbb{U}$, govern the robot's motions. The robot has access only to an {\bf approximate dynamics model} via a function $\dot{x} = \hat{f}_{\rho}(x,u)$ defined via a set of parameters $\rho$. The approximate model $\hat{f}$ is a simplification of $f$ and has a different expression, i.e.,  no choice of parameters $\rho$ will allow $\hat{f}$ to identify with $f$.  For instance, the robot assumes a flat, planar floor with known, uniform friction.  In reality, however, the workspace has: (i) different friction, which can be uniform or vary over the floor, and (ii) unmodeled traversable obstructions, such as speed bumps and ramps. Beyond friction, examples of parameters $\rho$ for a car-like robot include the steering and throttle gains.

A {\bf plan} $p(T)$ is a sequence of piece-wise constant controls $\{u^p_0, \ldots, u^p_{T-dt} \}$ of duration $T$, where each $u^p_t$ is executed for time $dt$. When $p(T)$ is executed at $x(t)$, it produces a {\bf trajectory}, i.e., a sequence of states $\tau_f( x(t), p(T) ) = \{x(t), \cdots, x(t+T)\}$ that respects the true model $f$, i.e.,: \vspace{-.1in}$$x(t'+i+dt) = \int^{t'+i+dt}_{t'+i} f( x(t), u^p_{i-1} ) \ dt.\vspace{-.1in}$$ 
Due to the model gap of the available model $\hat{f}_\rho$, the trajectory the robot follows $\tau_f( x(t), p(T) )$ does not match the planned trajectory $\tau_{\hat{f}_\rho}( x(t), p(T) )$ generated during the simulation process for the same plan $p(T)$.

The robot has access to {\bf noisy state estimates} $\hat{x}(t)$ given sensing.  
A {\bf controller} $\pi_{\hat{f}}( \hat{x}(t), \tau_{\hat{f}} )$ is employed to track the planned trajectory $\tau_{\hat{f}}$ given the noisy state observations $\hat{x}(t)$ and returns controls $u \in \mathbb{U}$. The controller aims to minimize an error $e_\tau$ between planning and the execution. Denote as $\tau_f( x(t), \pi_{\hat{f}})$ the trajectory executed by the robot of total duration $T$ when the controller $\pi_{\hat{f}}$ is employed to track the planned trajectory $\tau_{\hat{f}}$. A {\bf safe solution trajectory} $\tau_f(x_s,\pi_{\hat{f}})$ satisfies: (i) $\forall\ t \in [0,T]: \tau_f(x_s,\pi_{\hat{f}})(t) \in \mathbb{X}_\mathrm{f}$, and (ii) $\tau_f(x_s,\pi_{\hat{f}})(T) \in X_G$, i.e., all states upon execution are safe (collision-free) and lead to the goal region.

{\bf Problem Definition:} Given a start state $x_s \in \mathbb{X}_\mathrm{f}$, a goal region $X_G \subset \mathbb{X}_\mathrm{f}$, access to noisy state estimates $\hat{x}(t)$ and a controller $\pi_{\hat{f}}$ for tracking trajectories $\tau_{\hat{f}}$, compute plans $p(T)$ that result in safe solution trajectories $\tau_f(x_s,\pi_{\hat{f}})$.

\begin{figure}[t]
    \centering
    \begin{subfigure}{\linewidth}
    \includegraphics[width=0.89\linewidth]{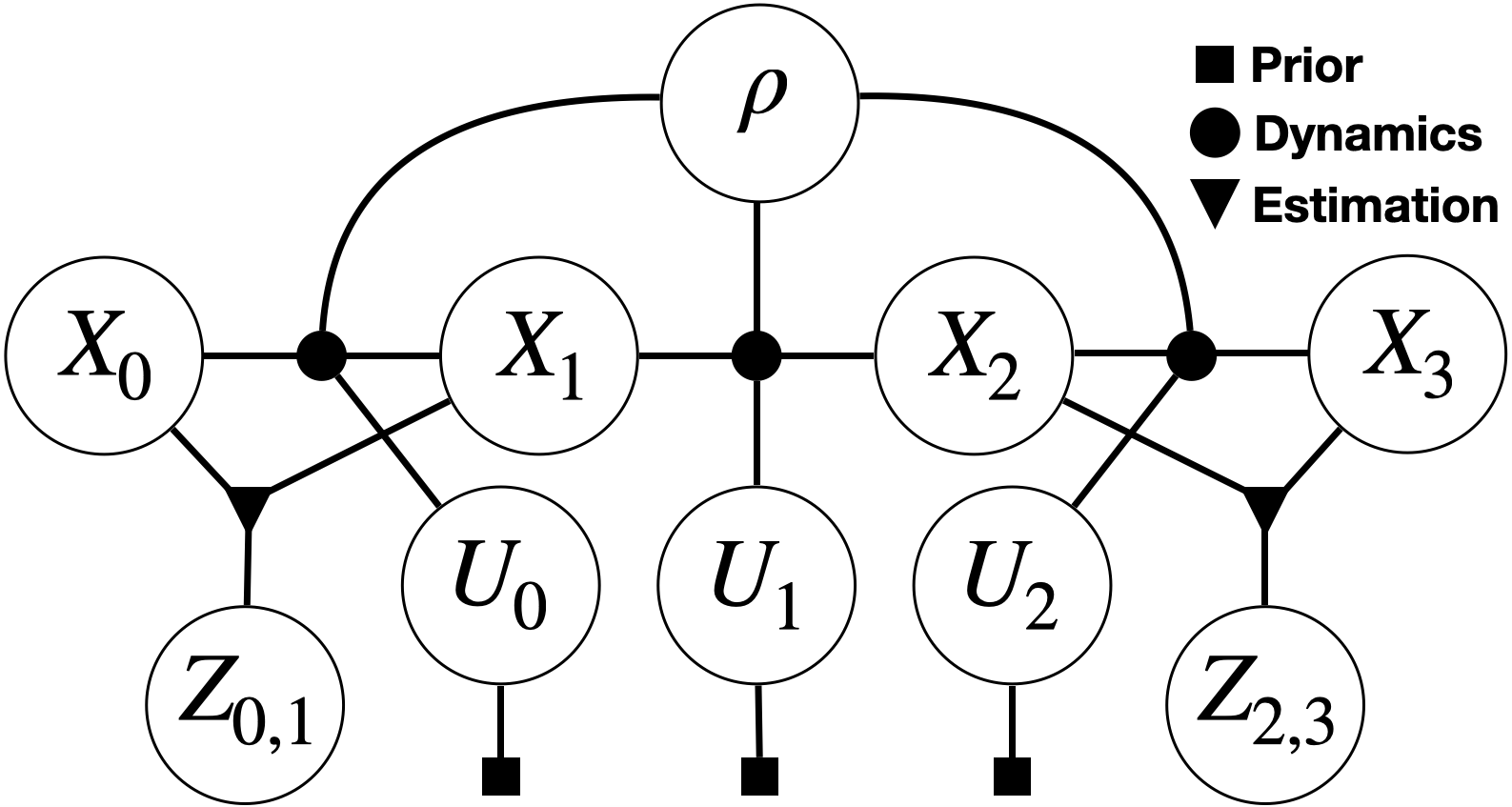}
    \end{subfigure} \\
    \caption{System identification using a factor graph for an observed trajectory $\{X_0, X_1, X_2, X_3\}$. Three types of factors are present: \textit{Prior} factors for the known applied controls of constant duration $\{U_0, U_1, U_2\}$; \textit{Dynamics} factors for the analytical model with unknown parameters $\rho$; and \textit{Estimation} factors for each observation $Z_{i,j}$ between states $X_i$ and $X_j$.}
    \vspace{-.25in}
    \label{fig:fg_sysid}
\end{figure}

Let ${\bf cost}(\tau)$ be the cost of an executed trajectory $\tau$. As a secondary objective, the objective is to also minimize the cost of the executed trajectory. In this work, the cost corresponds to trajectory duration. 

\noindent \textit{Other helpful notation: }A function $\mathbb{M}: \mathbb{X} \rightarrow \mathbb{Q}$ maps a state $x \in \mathbb{X}$ to its corresponding \textit{configuration space} point $q \in \mathbb{Q}$ ($q = \mathbb{M}(x)$). A distance function $d(\cdot,\cdot)$ is defined over $\mathbb{Q}$. In this work, the goal region is defined by a single configuration $q_G$ so that: $X_G = \{x \in \mathbb{X}_f \ \vert \ d(\mathbb{M}(x), q_G) < \epsilon \}$, or equivalently, $X_G = \mathcal{B}(q_G, \epsilon)$ where $\epsilon$ is a goal radius in $\mathbb{Q}$ according to function $d$. A heuristic $h: \mathbb{X} \rightarrow \mathbb{R}^+$ estimates the \textit{cost-to-go} of an input state $x$ to the goal region $X_G$.

\section{Proposed System and Methods}
\label{sec:proposed}

\subsection{Offline: System Identification via Factor Graphs}

The physical parameters $\rho$ of the analytical dynamical model $\hat{f}_{\rho}$ (e.g., steering offset, gains, etc.) are first identified given executed trajectories in a training environment. In particular, using observed trajectories $\tau$ of the system under a known plan $p(T)$, $\rho$ can be estimated by solving the following system identification problem:

\vspace{-.25in}
\begin{subequations}\label{eq:sysid}
\begin{align}
    \argmin_\rho \quad & || \tau_f - \tau_{\hat{f}} || \\
\textrm{s.t.} \quad & x(t+dt) = x(t) + \hat{f}_\rho(x(t), u)dt \label{eq:sysid_dynamics}\\
\quad & \hat{x}(t+\epsilon) = x(t+\epsilon) + N(0, \sigma) \label{eq:sysid_observation}
\vspace{-.3in}
\end{align}

The system identification is solved via least squares optimization on a factor graph (Fig.~\ref{fig:fg_sysid}). The dynamics factor implements equation~\ref{eq:sysid_dynamics} for a constant $dt$. The controls correspond to the executed plan and imposed via a prior factor. An asynchronous  observation $\hat{x}(t+\epsilon), \epsilon \in [0,1]$ assuming noise $\mathcal{N}(0,\sigma)$, is obtained between states $x(t)$ and $x(t+1)$. The estimation factor implements equation~\ref{eq:sysid_observation} by interpolating states $x(t)$ and $x(t+1)$ given $\epsilon$ to obtain $x(t+\epsilon)$. The initial guess is obtained by forward propagating $f_{\rho_0}(\hat{x}(0), u)$ for the duration of the plan.
\end{subequations}

\begin{figure*}
    \centering
    \includegraphics[width=\linewidth]{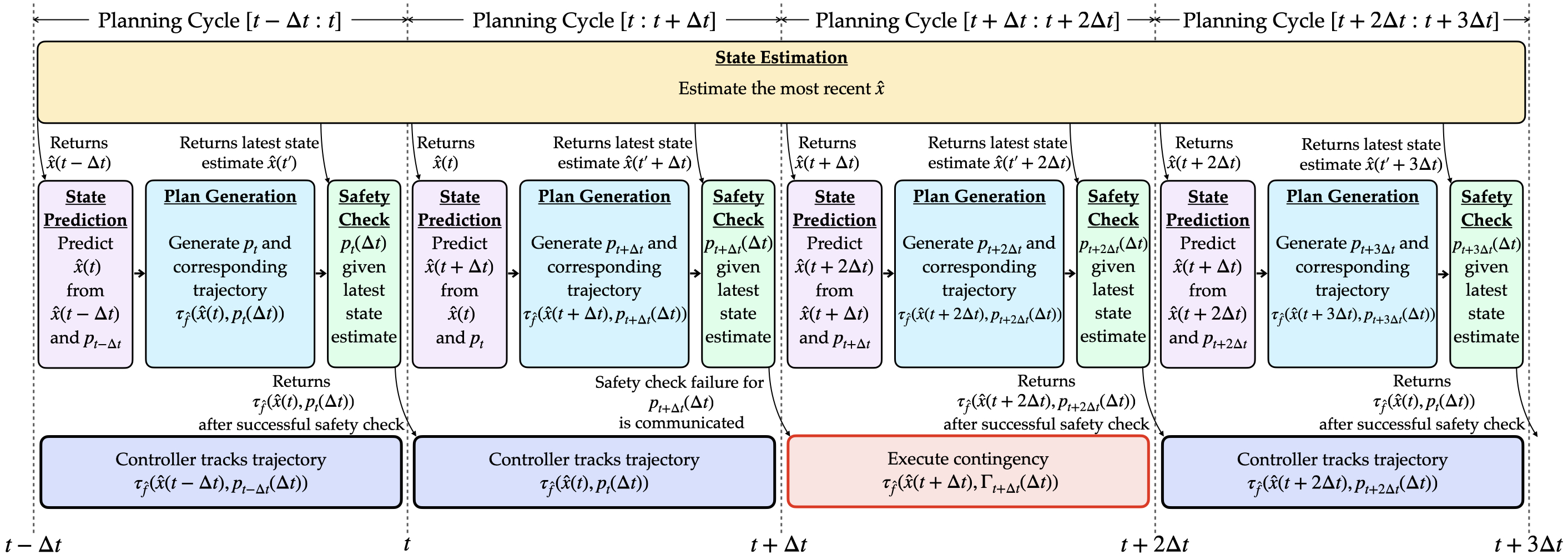}
    \vspace{-.2in}
    \caption{\name's online operation over different replanning cycles and integration with state estimation.}
    \vspace{-.25in}
    \label{fig:planning-cycles}
\end{figure*}

\subsection{Online: Safe Replanning Framework}

The sampling-based replanning framework considered in this work (Fig~\ref{fig:planning-cycles}) imposes a fixed planning cycle of duration $\Delta t$. For cycle $[t - \Delta t, t]$, the following steps are executed:

\begin{myitem}
    \item The current state $\hat{x}(t-\Delta t)$ is estimated given the most recent observations and the future robot state $\hat{x}(t)$, i.e., the initial state for the next planning cycle $[t, t + \Delta t]$, is predicted given the committed plan $p_{t-\Delta t}$ and model $\hat{f}_{\rho}$.
    \item Then, an SBMP generates a (long-horizon) plan $p_t$, whose initial $\Delta t$ duration produces the trajectory $\tau_{\hat{f}}(\hat{x}(t),p_t(\Delta t))$.
    \item Right before the completion of cycle $[t - \Delta t, t]$, a safety check is performed given the chosen plan $p_t$ and the latest state estimate for $\hat{x}(t)$. If the trajectory $\tau_{\hat{f}}(\hat{x}(t),p_t(\Delta t))$ is still deemed safe, it is communicated to the low-level controller. If not, a trajectory corresponding to a contingency maneuver $\Gamma_t(\Delta t)$ is communicated instead. 
    \item The communicated trajectory will be tracked by a feedback controller during the following planning cycle $[t, t + \Delta t]$.
\end{myitem}

Algo.~\ref{alg:tree-sbmp} outlines the high-level operation of the sampling-based tree motion planner (Tree-SBMP) adopted. In every cycle $[t - \Delta t, t]$, the planner incrementally updates a tree data structure of states reachable from the initial state of the next planning cycle $\hat{x}(t)$. It contains (a) a \emph{retainment} step that reuses information from the long-horizon plan from the previous cycle $p_{t-\Delta t}$; (b) a \emph{tree expansion} step that is informed and aims to return high-quality solutions quickly; and (c) a \emph{safety check} step that takes into account the maximum possible deviation between the estimated robot state $\hat{x}(t)$ and the true state $x(t)$.

\vspace{-.2in}
\begin{algorithm}[h!]
\SetAlgoLined
{\color{blue} // {\tt Retainment}}  \\
Set $\texttt{TREE}.root = \hat{x}(t), best \leftarrow \emptyset, best\_cost \leftarrow \infty;$ \\
$\tau_{prev}(\hat{x}(t),p_{t-\Delta t}) \leftarrow 
\int \hat{f}( \hat{x}(t),p_{t - \Delta t});$ \\
\If{$\tau_{prev}.\text{length} < \Delta t$} 
{
Add collision-free subset of $\tau_{prev}$ to $\texttt{TREE}$ ;
}
{\color{blue} // {\tt Safety check for retained plan}}  \\
\ElseIf{$\tau_{prev} \in \mathbb{X}_f$ \& {\tt SAFETY}($\tau_{prev}[t + \Delta t], \Gamma$)}
{
Add $\tau_{prev}$ to $\texttt{TREE}$;
}
{\color{blue} // {\tt Tree Expansion}}  \\
\While{ $planning\_time$ \textrm{has not been reached}}
{
$x_\text{sel}(t') \leftarrow$ Select Node from \texttt{TREE} ($t' \geq t)$; \\ 
Select plan $p_\text{cand} = (u, dt)$ to expand from $x_\text{sel}$; \\
$\tau_\text{cand}(x_\text{sel}(t'),p_\text{cand}) \leftarrow \int \hat{f}(x_\text{sel}(t'),p_\text{cand})$; \\
$added \leftarrow false;$ \\
\If{$t' + dt < \Delta t$ \& $\tau_\text{cand} \in \mathbb{X}_f$}
{
Add $\tau_{cand}$ to $\texttt{TREE}$; $added \leftarrow true;$
}
{\color{blue} // {\tt Safety check for new edge}}  \\
\ElseIf{$\tau_{cand} \in \mathbb{X}_f$ \& {\tt SAFETY}($\tau_{cand}[t + \Delta t], \Gamma$)}
{
Add $\tau_{cand}$ to $\texttt{TREE}$; $added \leftarrow true;$
}
\If{added \& $\tau_\text{cand}.\text{end}() \in X_G$ \& $\tau_\text{cand}.\text{length}() < best\_cost$}
{
$best \leftarrow \tau_\text{cand}.\text{end}();$ \\ 
$best\_cost \leftarrow \tau_\text{cand}.\text{length}();$
}
}
\If{$best\_cost == \infty$}
{
$best \leftarrow \argmin_\texttt{TREE} h(x);$ \\
}
\Return {\text{trajectory on tree leading to} $best$ state;}
\caption{Tree-SBMP\\ 
Inputs: $\hat{x}(t),\ \hat{f},\ p_{t-\Delta t},\ \mathbb{X}_f,\ X_G,\ \Gamma,\ planning\_time, h$}
\label{alg:tree-sbmp}
\end{algorithm}
\vspace{-.2in}

\noindent {\bf Retainment:} \name\ uses a longer planning horizon relative to standard MPC approaches. As a result, the plan computed during the previous planning cycle may still contain useful guidance for returning a new solution given the latest state estimation $\hat{x}(t)$. If the previous plan $p_{t-\Delta t}$ has controls beyond $t$, then the subset of the plan beyond $t$ is forward propagated from the root of the new tree, i.e., $\hat{x}(t)$, to obtain the trajectory $\tau_{prev}( \hat{x}(t), p_{t-\Delta t} )$. The subset of the trajectory from $t$ is retained for the current planning cycle.


\noindent {\bf Expansion:} Once the feasible and safe subset of the previous solution is retained, then the Tree-SBMP algorithm further expands the tree given the available $planning\_time$. It selects an existing tree node/state $x_\text{sel}$ to expand, which will occur at time $t'> t$. Then, it generates a control sequence $(u, dt)$ and propagates it from $x_\text{sel}$ resulting in a candidate trajectory $\tau_{cand}$. If the edge is deemed collision-free and safe, it is added to the tree. If the tree discovers states in $X_G$, the best-found solution according to $cost$ is returned. If no state inside $X_G$ is generated, the solution that terminates at the state with the best heuristic cost-to-go $h(x)$ is returned.

\noindent {\bf Safety Check:} Returning a collision-free plan for the next planning cycle does not guarantee safety for second-order systems even in a static environment and for a perfect dynamics model due to Inevitable Collision States (ICS) \cite{fraichard2004inevitable, bekris2007greedy}. To deal with ICS in static environments and given a perfect model, it is possible to use \textit{contingency plans} $\Gamma$ (e.g., braking maneuvers) and impose the following invariant: At the end of the cycle, i.e., at a state $x[t + \Delta t]$, the robot should be at a safe state, i.e., $\exists\ \gamma \in \Gamma \ \text{s.t.}\ \tau_{f}(x[t + \Delta t],\gamma)$ is collision-free until the robot comes to a stop. 


\vspace{-.2in}
\begin{algorithm}
\SetAlgoLined
\For{$\gamma \in \Gamma$}
{
$\tau_{safe}( \hat{x}, \gamma ) \leftarrow  \int \hat{f}( \hat{x}(t), \gamma );$ \\
$check \leftarrow true;$ \\
\For{all $x_\text{safe} \in \tau_\text{safe}$ \& while $check$ is true}
{
\If{{\tt DistToClosestObst}$(x_\text{safe}) < \delta$}
{
$check \leftarrow false;$\\
}
}
\If{check}
{
\Return{true};
}
}
\Return{false};
\caption{SAFETY( $\hat{x}, \hat{f}, \Gamma$ )}
\label{alg:SAFETY}
\end{algorithm}
\vspace{-.2in}

This work extends the safety notion given an approximate model $\hat{f}$ and noise in state estimation. At time $t$ there is perception error regarding the estimate $\hat{x}(t)$ relative to the true state $x(t)$. The execution of the plan $p_t(\Delta t)$ and of a potential contingency $\gamma$ of duration $\gamma.t$ after it, will also result in execution errors during the time frame $[t+\Delta t + \gamma.t]$, when the concatenation $p_t(\Delta t)|\gamma$ of plans is executed. This work {\bf assumes} that the combination of these errors results in deviations between the predicted $\tau_{\hat{f}}(\hat{x},p_t(\Delta t)|\gamma)$ and the true $\tau_{f}(x,p_t(\Delta t)|\gamma)$, which are upper bounded by a distance $\delta$ given the robot's dynamics and the available perception. 

Given this assumption, if there is a plan $p_t$ and a contingency $\gamma \in \Gamma$ after it so that the predicted trajectory $\tau_{\hat{f}}(\hat{x}(t),p_t(\Delta_t)|\gamma)$ maintains a $\delta$-clearance from the obstacles until the robot comes to a complete stop, then $\hat{x}(t)$ is deemed safe. Lines 7-8 and 18-19 in Alg. 1 perform this safety check. For every state on the tree that will occur at time $[t + \Delta t]$, and which is a candidate initial state for the consecutive cycle, they call the {\tt SAFETY} function (Alg. 2), which evaluates whether contingency plans are guaranteed to provide $\delta$-clearance from the obstacles out of these states. The same requirement is imposed for plans on the tree within the initial $\Delta t$ cycle. 

\begin{theorem}
Assume a 2$^{nd}$-order system executing replanning with {\tt Tree-SBMP} in a static environment where deviations between predicted and true trajectories are upper bounded by distance $\delta$. For safety, it is sufficient to compute plans $p_t(\Delta)$ followed by braking maneuvers $\gamma$ so that the resulting trajectory $\tau_{\hat{f}}(\hat{x}(t),p_t(\Delta_t)|\gamma)$ maintains $\delta$-clearance with obstacles.
\end{theorem}
\vspace{-1mm}
\begin{proof}
Assume the robot is safe at time $t$ and has computed a plan that satisfies the assumptions.  The robot will not collide if it executes $p_t(\Delta t)$ since in the worst case it will deviate by distance $\delta$ from its predictions but the predicted trajectory $\tau_{\hat{f}}(\hat{x}(t),p_t(\Delta_t))$  is at least $\delta$ distance away from obstacles. There are two cases during the next cycle $[t, t + \Delta t]$: (a) The planner produces a new safe plan and contingency for the next period $t + \Delta t$, thus maintaining the invariant. (b) If the planner fails to compute a safe plan, the contingency $\gamma \in \Gamma$ can be executed at $t + \Delta t$, bringing the robot to a collision-free stop under the assumption. So, in every case, there is a collision-free plan for the future.
\end{proof}



\noindent \textbf{Feedback Control.} The above framework has been integrated in the experiments with different low-level controllers. The na\"ive baseline is {\tt Open-Loop}, which blindly executes the planners' solution. Two closed-loop solutions use the latest state estimates $\hat{x}(t)$: ({\tt Geometric}) {\it path following} finds the closest configuration $q_\text{near}$ on the solution trajectory (within a window of the previous closest point) to $\hat{x}(t)$. It uses a PID controller to navigate the robot towards a lookahead point that is a fixed length away from $q_\text{near}$. ({\tt Kinodynamic}) \textit{trajectory tracking}  uses the Stanley controller \cite{hoffmann2007autonomous} to track the planned trajectory given the dynamics model $\hat{f}_\rho$. All controllers are unaware of obstacles. Both closed-loop controllers are unaware of the robot's actuation limits, and may return controls that need to be clamped before sending them to the robot's onboard controller. 


\begin{table*}[ht!]
\vspace{0.05in}
\centering
\begin{tabular}{|c||cccc||cccc|}
\hline
                        & \multicolumn{4}{c||}{\tt Turns}                                                                     & \multicolumn{4}{c|}{\tt Boxes}                                                                     \\ \hline
Planning + Control Framework                 & \multicolumn{1}{c|}{\tt Succ} & \multicolumn{1}{c|}{\tt Coll} & \multicolumn{1}{c|}{\tt Timeout} & $T_\text{ex}$ (s) & \multicolumn{1}{c|}{\tt Succ} & \multicolumn{1}{c|}{\tt Coll} & \multicolumn{1}{c|}{\tt Timeout} & $T_\text{ex}$ (s) \\ \hline
{\tt OneShot + OpenLoop}            & \multicolumn{1}{c|}{0}     & \multicolumn{1}{c|}{30}     & \multicolumn{1}{c|}{0}        &    N.A.     & \multicolumn{1}{c|}{0}     & \multicolumn{1}{c|}{30}     & \multicolumn{1}{c|}{0}        &    N.A.     \\ \hline
{\tt OneShot + Geometric}                 & \multicolumn{1}{c|}{19}     & \multicolumn{1}{c|}{11}     & \multicolumn{1}{c|}{0}        &     16.88    & \multicolumn{1}{c|}{8}     & \multicolumn{1}{c|}{22}     & \multicolumn{1}{c|}{0}        &     11.49    \\ \hline
{\tt OneShot + Kinodynamic}              & \multicolumn{1}{c|}{23}     & \multicolumn{1}{c|}{7}     & \multicolumn{1}{c|}{0}        &     15.86    & \multicolumn{1}{c|}{17}     & \multicolumn{1}{c|}{13}     & \multicolumn{1}{c|}{0}        &    13.81     \\ \hline
{\tt Replanner + OpenLoop} & \multicolumn{1}{c|}{5}     & \multicolumn{1}{c|}{25}     & \multicolumn{1}{c|}{0}        &  17.5     & \multicolumn{1}{c|}{0}     & \multicolumn{1}{c|}{30}     & \multicolumn{1}{c|}{0}        &  N.A.       \\ \hline
{\tt Cons. Replanner + OpenLoop}   & \multicolumn{1}{c|}{11}     & \multicolumn{1}{c|}{15}     & \multicolumn{1}{c|}{4}        &     30.5    & \multicolumn{1}{c|}{8}     & \multicolumn{1}{c|}{13}     & \multicolumn{1}{c|}{9}        &    39.33\\ \hline
\name & \multicolumn{1}{c|}{27}     & \multicolumn{1}{c|}{3}     & \multicolumn{1}{c|}{0}        &       16.65  & \multicolumn{1}{c|}{23}     & \multicolumn{1}{c|}{7}     & \multicolumn{1}{c|}{0}        &  17.39       \\ \hline
{\tt Conservative} \name   & \multicolumn{1}{c|}{30}     & \multicolumn{1}{c|}{0}     & \multicolumn{1}{c|}{0}        &   18.9      & \multicolumn{1}{c|}{25}     & \multicolumn{1}{c|}{1}     & \multicolumn{1}{c|}{4}        &     40.3    \\ \hline
\end{tabular}
\caption{Evaluation on simulation environments with similar physical properties as the tuned planning model. Each method is executed 30 times.}
\vspace{-.2in}
\label{tab:easy-environments}
\end{table*}

\section{Experiments}
\label{sec:experiments}

The \textbf{mobile robot system} considered in the experimental evaluation is the low-cost, open-source MuSHR racing platform \cite{srinivasa2019mushr}. The controls are $[\nu, \phi]$, where $\nu$ is throttle, and $\phi$ is the desired steering angle. The \textbf{planning model} has a state space of $[x,y,\theta,v]$, where $(x,y,\theta) \in {\tt SE(2)}$ is the pose of the car in the world frame, and $v \in [v_\text{min}, v_\text{max}]$ is forward velocity. For integration purposes, a 4-th order Runge Kutta approach is used. The model's parameters correspond to $[L, \phi_\text{diff}, v_\delta]$, i.e., the wheelbase length, steering angle offset, and the throttle gain.

\begin{figure}[ht!]
    \vspace{-.1in}
    \centering
    \vspace{1mm}
    \begin{subfigure}{\linewidth}
    \includegraphics[width=0.49\linewidth]{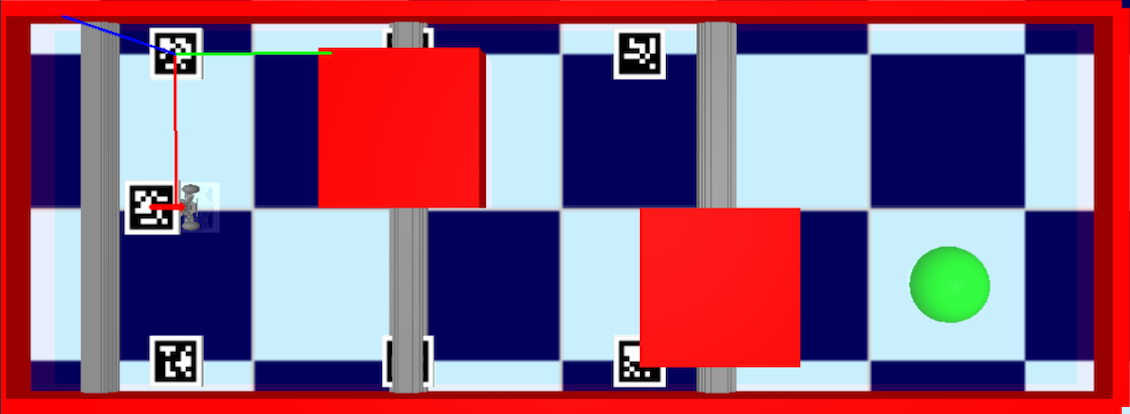}
    \includegraphics[width=0.49\linewidth]{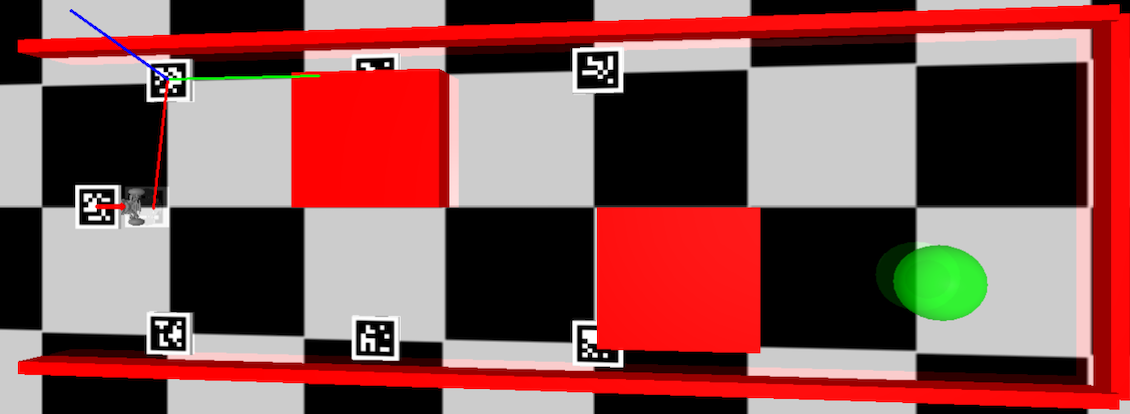}
    \end{subfigure} \\
    \vspace{1mm}
    \begin{subfigure}{\linewidth}
    \includegraphics[width=0.49\linewidth]{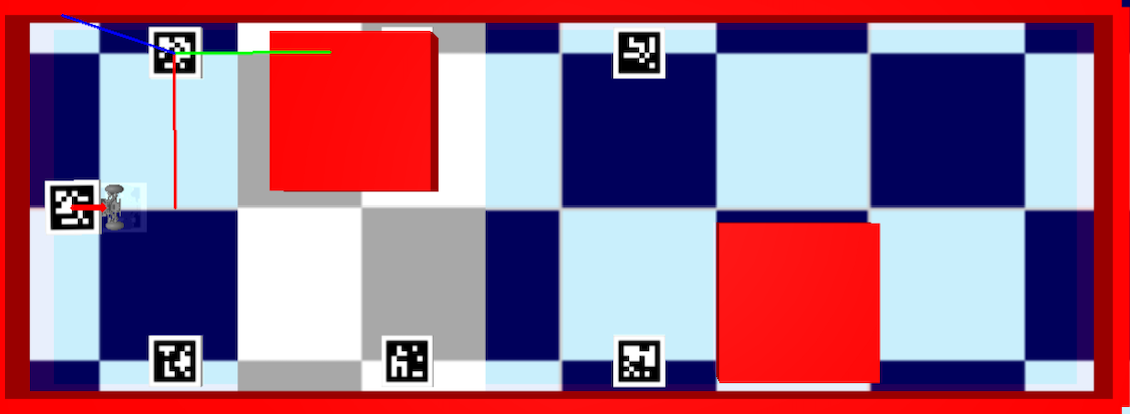}
    \includegraphics[width=0.49\linewidth]{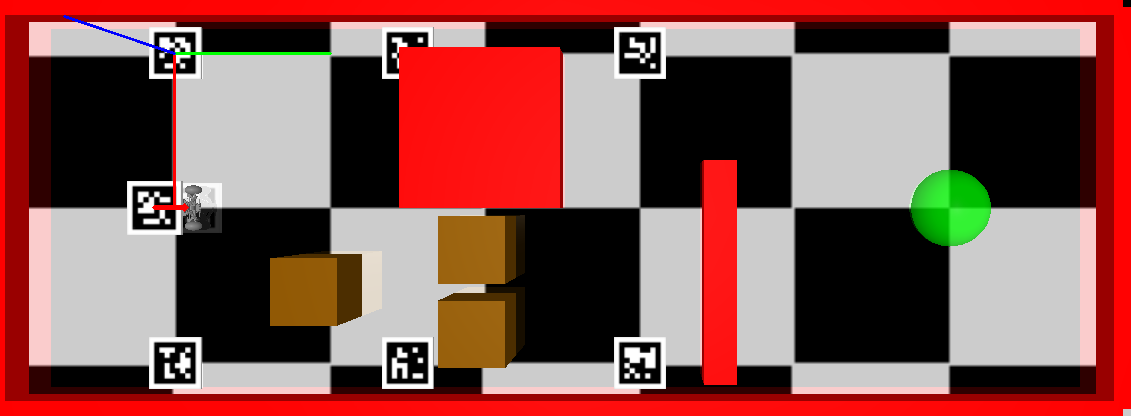}
    \end{subfigure} \\
    \vspace{-0.05in}
    \caption{Environments with features not modeled by the planner. Goal set  shown in green. Top: (L-R) {\tt Bump}, {\tt Slope}. Btm: (L-R) {\tt Slip}, {\tt Movable.}}
    \label{fig:hard-benchmarks}
    \vspace{-.2in}
\end{figure}

For evaluation in simulation, the \textbf{ground-truth} system is modeled in MuJoCo \cite{todorov2012mujoco}. ArUco tags \cite{garrido2016generation, romero2018speeded} are used to detect the robot's location with noise. For state estimation purposes, the robot's current velocity is the commanded desired velocity from the previous timestep. Two types of \textbf{challenges} are considered for evaluation in simulation. In two environments, {\tt Turns} and {\tt Boxes}, the approximate model $f_\rho$ has been identified in an environment with the same physical properties. Thus, \name\ must only deal with sensor noise and execution error. In the second set of challenges (Fig. \ref{fig:hard-benchmarks}), \name\ must also deal with unmodeled aspects, such as slopes, uneven terrain, and movable, lightweight obstacles. In all environments, the robot must not collide with fixed static obstacles in the scene.

The performance is measured according to these \textbf{metrics}: (1) $\#$ of trials where the robot navigated to the goal without collision ({\tt Succ}), (2) $\#$ of trials where the robot has an unrecoverable collision ({\tt Coll}), (3) $\#$ of trials where the robot could not reach the goal within a preset timeout limit ({\tt Timeout}),  and (4) Average execution duration across the successful trials $T_\text{ex}$. Each method is run 30 times on each benchmark to account for different random seeds.


{\bf Baselines:} The planner used in the experiments is the AO Dominance-Informed Region Tree (DIRT) \cite{LB-DIRT}.  The {\tt OneShot} baseline calls DIRT for a solution and passes the trajectory to the controller without replanning. The {\tt Replanner} uses the proposed framework to frequently update the trajectory but does not use a feedback controller. The proposed method, \name, uses the replanning framework and the {\tt Kinodynamic} trajectory tracker.

{\bf Evaluation in Simulation - Tuned Model:} Table~\ref{tab:easy-environments} reports results on the {\tt Turns} and {\tt Boxes} environments.

\noindent \textbf{Q1.} \textit{Given single-shot planning with an approximate model, what is the safest control strategy?} While the planning model has been tuned with the same physical properties as {\tt Turns} and {\tt Boxes}, open-loop execution of the plan results in a collision as small deviations compound the error over time. The {\tt Geometric} controller often tracks the path in {\tt Turns} environment (19/30 Succ) but fails more frequently in the {\tt Boxes} environment. The {\tt Kinodynamic} trajectory follower is more consistent than the {\tt Geometric} path follower on average but does not fully address the gap.

\noindent \textbf{Q2.} \textit{Does replanning improve success rate?} The {\tt Replanner} enables the robot to reach the goal in more trials than the {\tt OneShot} planner in open loop execution.  \name, which also performs replanning, is the best-performing strategy. 

\noindent \textbf{Q3.} \textit{What is the effect of the proposed safety checks?} {\tt Conservative} \name\ implements the safety framework of Algorithm~\ref{alg:SAFETY}. Similarly for the {\tt Cons. Replanner + OpenLoop} baseline. The application of the safety framework appears to increase the success rate but at the same time results in longer duration trajectories as at multiple points during execution, the robot selects to revert to a braking maneuver. The single trial that resulted in collision in the {\tt Boxes} environment for {\tt Conservative} \name\ arose from a communication delay between the replanner triggering the contingency plan, and the simulator environment executing it. When the safety radius $\delta$ is further increased to deal with this issue, more timeouts are observed as the robot becomes increasingly conservative close to obstacles. Exploring how to dynamically adapt the safety radius $\delta$ is an interesting future direction.


{\bf Evaluation in Simulation - Unmodeled Features:} The following experiments focus on the environments of Fig~\ref{fig:hard-benchmarks}, which include physical attributes not captured by the planning model. Four different combinations of replanning and control are evaluated. 

\begin{table}[h!]
\vspace{0.1in}
\centering
\begin{tabular}{|c|c|c|c|c|}
\hline
Framework & {\tt Succ} & {\tt Coll} & {\tt Timeout} & $T_\text{ex}$ (s) \\ \hline
{\tt OneShot+Open}&  0    &  30    &   0     &    N.A.    \\ \hline
{\tt OneShot+Geometric}             &  0    &   27   &    3     &      N.A.  \\ \hline
{\tt OneShot+Kinodynamic}&   4   &   16   &   10      &    23.18    \\ \hline
{\tt KRAFT}            &  19    &   11   &     0    &    29.94     \\ \hline
\end{tabular}
\vspace{-.05in}
\caption{Evaluation on the {\tt Slope} environment.}
\label{tab:slope-results}
\vspace{-.2in}
\end{table}

\begin{table*}[t]
\vspace{0.05in}
\centering
\begin{tabular}{|c||cccc||cccc|}
\hline
                        & \multicolumn{4}{c||}{{\tt Turns-Real}}                                                                     & \multicolumn{4}{c|}{{\tt Boxes-Real}}                                                                     \\ \hline
                        
Planning + Control Framework                 & \multicolumn{1}{c|}{\tt Succ} & \multicolumn{1}{c|}{\tt Coll} & \multicolumn{1}{c|}{\tt Timeout} & $T_\text{ex}$ (s) & \multicolumn{1}{c|}{\tt Succ} & \multicolumn{1}{c|}{\tt Coll} & \multicolumn{1}{c|}{\tt Timeout} & $T_\text{ex}$ (s) \\ \hline

{\tt OneShot + OpenLoop}            & \multicolumn{1}{c|}{0}     & \multicolumn{1}{c|}{10}     & \multicolumn{1}{c|}{0}        &    N.A.     & \multicolumn{1}{c|}{0}     & \multicolumn{1}{c|}{7}     & \multicolumn{1}{c|}{3}        &    N.A.     \\ \hline

{\tt OneShot + Kinodynamic}& \multicolumn{1}{c|}{2}     & \multicolumn{1}{c|}{8}     & \multicolumn{1}{c|}{0}        &  15.325& \multicolumn{1}{c|}{4}     & \multicolumn{1}{c|}{6}     & \multicolumn{1}{c|}{0}        &  25.90\\ \hline

{\tt Replanner + OpenLoop}& \multicolumn{1}{c|}{1}     & \multicolumn{1}{c|}{9}     & \multicolumn{1}{c|}{0}        &     13.26& \multicolumn{1}{c|}{5}     & \multicolumn{1}{c|}{5}     & \multicolumn{1}{c|}{0}        &     22.95\\ \hline

{\tt Cons. Replanner + OpenLoop}& \multicolumn{1}{c|}{7}     & \multicolumn{1}{c|}{3}     & \multicolumn{1}{c|}{0}        &     20.61& \multicolumn{1}{c|}{8}     & \multicolumn{1}{c|}{2}     & \multicolumn{1}{c|}{0}        &    31.75\\ \hline

\name& \multicolumn{1}{c|}{10}     & \multicolumn{1}{c|}{0}     & \multicolumn{1}{c|}{0}        &     14.06& \multicolumn{1}{c|}{8}     & \multicolumn{1}{c|}{2}     & \multicolumn{1}{c|}{0}        &    26.91\\ \hline

{\tt Conservative} \name & \multicolumn{1}{c|}{10}     & \multicolumn{1}{c|}{0}     & \multicolumn{1}{c|}{0}        &       31.58& \multicolumn{1}{c|}{9}     & \multicolumn{1}{c|}{1}     & \multicolumn{1}{c|}{0}        &  45.55\\ \hline
\end{tabular}
\caption{Real-world evaluation on an environment where the planning model has been tuned. Each method is executed 10 times. }
\label{tab:real-environments}
\vspace{-.25in}
\end{table*}

In the {\tt Slope} environment (Table~\ref{tab:slope-results}), the floor is at a slope of \ang{7.16} not known to the planner. Moreover, the state estimation process is also unaware of the slope, which introduces noise in state estimates. Consequently, the {\tt OneShot} trajectories cannot be reliably tracked even with the {\tt Geometric} or {\tt Kinodynamic} controllers. Replanning in \name\ helps to improve success rate.

\begin{table}[!htbp]
\vspace{-.1in}
\centering
\begin{tabular}{|c|c|c|c|c|}
\hline
Framework & {\tt Succ} & {\tt Coll} & {\tt Timeout} & $T_\text{ex}$ (s) \\ \hline
{\tt OneShot+Open}         &  0    &  30    &  0      &     N.A.   \\ \hline
{\tt OneShot+Geometric}            &  6    &   24   &    0     &   16.86     \\ \hline
{\tt OneShot+Kinodynamic}        &   8   &   16   &   6     &    14.45    \\ \hline
\name            &  20    &   10   &     0    &    16.4    \\ \hline
\end{tabular}
\vspace{-.05in}
\caption{Evaluation on the {\tt Slip} environment.}
\label{tab:slip-results}
\vspace{-.15in}
\end{table}

The {\tt Slip} environment is similar to the {\tt Turns} environment but exhibits different friction coefficients over a portion of the floor. This leads to the robot slipping, which is not predicted by the planning model. The feedback controllers tend to command the robot to move at lower velocities, which helps in a few cases to track the planned trajectory. Replanning in \name\ again helps to improve success rate.

\begin{table}[!htbp]
\vspace{-.1in}
\centering
\begin{tabular}{|c|c|c|c|c|}
\hline
Framework & {\tt Succ} & {\tt Coll} & {\tt Timeout} & $T_\text{ex}$ (s) \\ \hline
{\tt OneShot+Open}        &   0   &  30    &   0     &   N.A.     \\ \hline
{\tt OneShot+Geometric}           &   13   &    16  &     1    &    17.95    \\ \hline
{\tt OneShot+Kinodynamic}       &   17   &   4    &  9       &     16.17    \\ \hline
\name           &  24    &  6    &  0       &    19.33     \\ \hline
\end{tabular}
\vspace{-.05in}
\caption{Evaluation on the {\tt Bump} environment.}
\label{tab:bump-results}
\vspace{-.15in}
\end{table}

The {\tt Bump} environment contains speed bumps that impede the robot's progress. Occasionally, the {\tt Geometric} controller makes progress by applying a higher throttle command proportional to the distance to the next target waypoint. The {\tt Kinodynamic} controller provides a higher throttle input to reduce tracking error, allowing it to navigate the bumps more frequently. Integrating with replanning in \name \ provides trajectories that are ahead of the robot with a higher velocity (based on its prediction of future states using the planning model) that further increase success rate.


\begin{table}[!htbp]
\vspace{-.1in}
\centering
\begin{tabular}{|c|c|c|c|c|}
\hline
Framework & {\tt Succ} & {\tt Coll} & {\tt Timeout} & $T_\text{ex}$ (s) \\ \hline
{\tt OneShot+Open}         &  0    &  30    &  0      &    N.A.   \\ \hline
{\tt OneShot+Geometric}             &  1    &  28    &  1       &     24.96  \\ \hline
{\tt OneShot+Kinodynamic}         &   4   &   24   &   2     &    18.79    \\ \hline
\name             &  17    &   12   &     1    &    29.32    \\ \hline
\end{tabular}
\vspace{-.05in}
\caption{Evaluation on the {\tt Movable} environment.}
\label{tab:movable-results}
\vspace{-.15in}
\end{table}

In the {\tt Movable} benchmark, the environment contains pushable boxes that the planner is unaware of. Even with control feedback, the {\tt OneShot} approaches cannot deal with the difference between planned and executed trajectories. The {\tt Replanner + Kinodynamic} integration in \name\ is again more successful.

{\bf Evaluation on the Real Platform:} The proposed framework has been tested with a real MuSHR platform in terms of safely navigating away from obstacles towards a goal region. The pose of the robot is tracked using ArUco markers by two different cameras. The parameters of the analytical planning model for the MuSHR are estimated via the factor graph estimation approach given a few nominal open-loop plans on the same floor but without obstacles. The max. throttle applied to the robot is slightly adapted to ensure smoother tracking of its pose.

\begin{figure}[h!]
\vspace{-.1in}
    \centering
    \begin{subfigure}{\linewidth}
    \centering
    \includegraphics[width=0.49\linewidth]{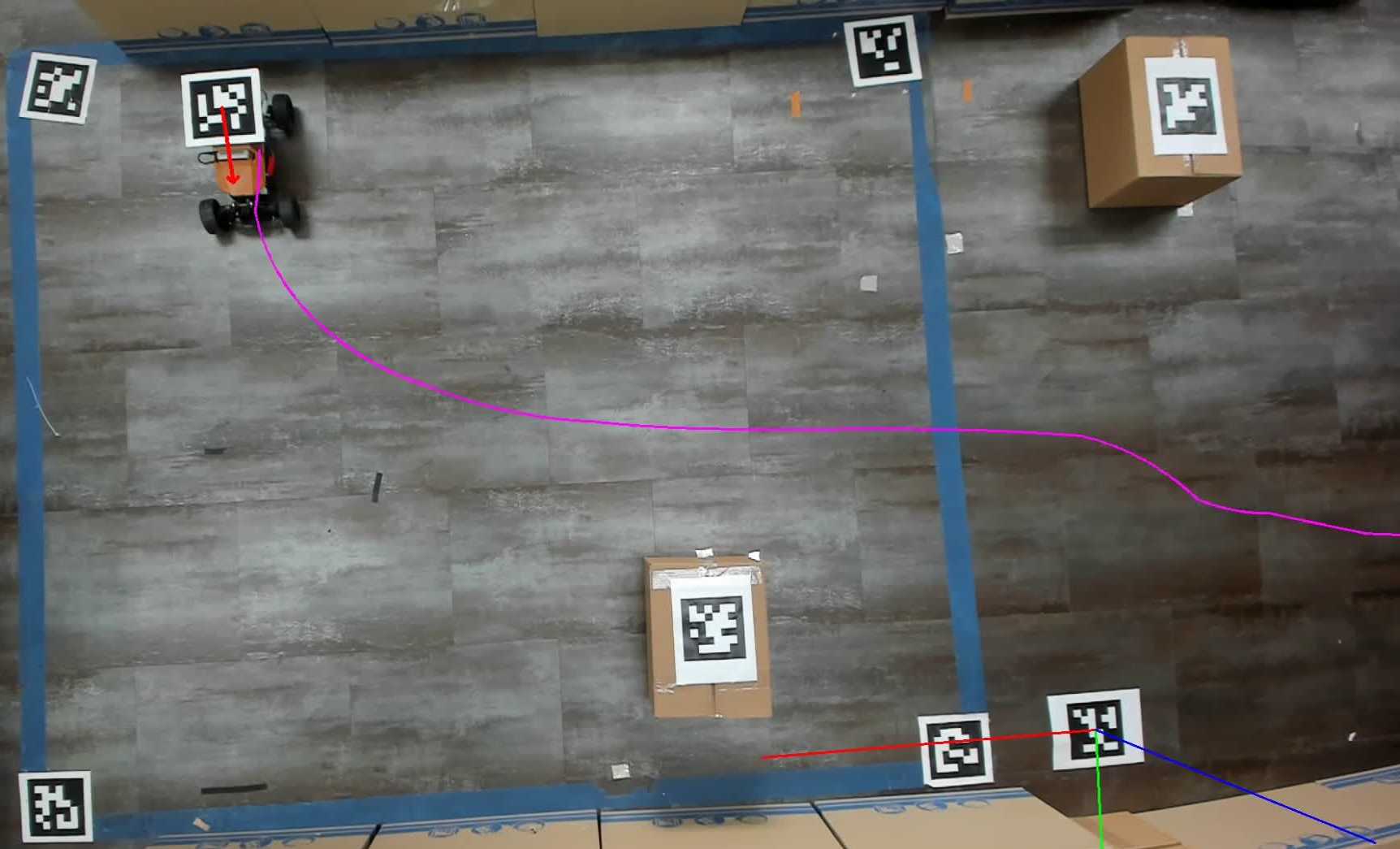}
    \includegraphics[width=0.49\linewidth]{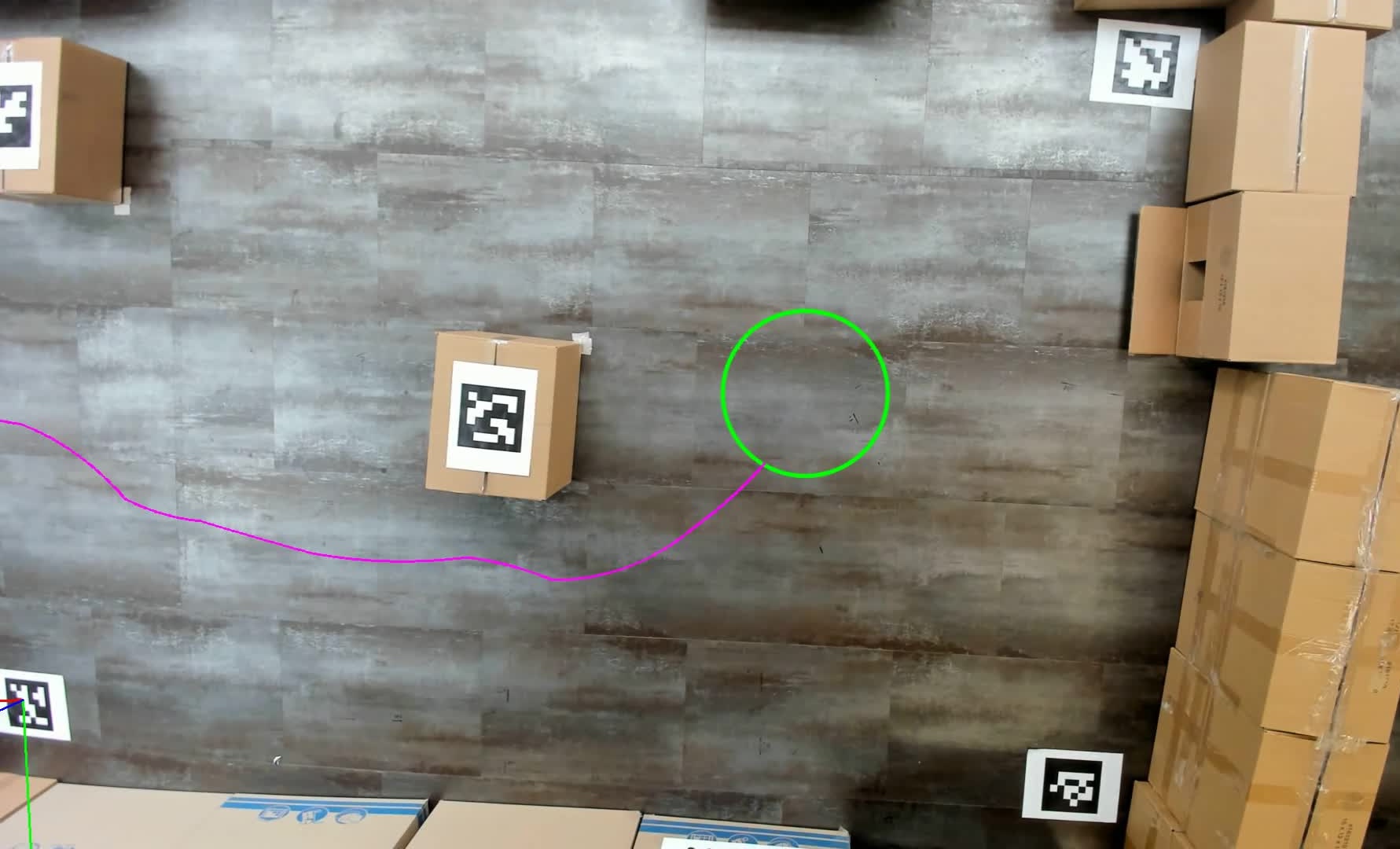}
    \end{subfigure} \\
    \vspace{-.05in}
    \caption{The {\tt Boxes-Real} environment as seen from the two cameras, which have some small field of view overlap in the middle. The robot's initial state is on the left side, while the desired goal state (green circle) is on the right. The planned trajectory is shown in purple.}
    \label{fig:real-world-benchmarks}
\vspace{-.2in}
\end{figure}

Two environments were tested: {\tt Turns-Real} (Fig.~\ref{fig:intro-figure}) and {\tt Boxes-Real} (Fig.~\ref{fig:real-world-benchmarks}), similar to the simulated ones. Table~\ref{tab:real-environments} evaluates the performance on these environments using the same metrics and planning and control frameworks as in Table~\ref{tab:easy-environments}. The results are consistent with the trends observed in simulation. \name\ consistently allows the robot to safely navigate to the goal. {\tt Conservative} \name\ is slightly more successful at the cost of higher execution duration, due to frequent use of contingencies.

On the real platform, significantly larger model gap and communication delays were observed. Again a delay in the communication of the contingency plan was noted during a single failure trial of the {\tt Conservative} \name. Future work will investigate tighter integration of perception and control, either via onboard sensing or executing the replanning framework on the robot's onboard computer to minimize communication overhead.


\section{Discussion}
\label{sec:conclusion}

This paper has presented \name, a framework aimed at enhancing the safety and efficiency of trajectory execution for robots operating with non-trivial dynamics in partially modeled environments. 
\name\ integrates kinodynamic replanning with a trajectory follower, which allows the approach to provide feedback on state estimation updates at multiple levels and in this way improve the robot's ability to navigate despite model inaccuracies and noise.

The accompanying experiments demonstrate that while \name\ significantly outperforms the alternatives, it can still face challenges when aspects of the environment significantly deviate from the available model. Future work will focus on further improving the ability of such methods to address such challenges by dynamically adapting the model over the planner as they start observing deviations.  Developing contingency plans tailored to specific scenarios will also be crucial for improving the proposed framework's safety as well as the efficiency of the conservative version of \name. 

\bibliographystyle{format/IEEEtran}
\bibliography{refs.bib}

\end{document}